\PassOptionsToPackage{table,xcdraw}{xcolor}
\documentclass[11pt,letterpaper]{article}
\usepackage[margin=1in]{geometry}
\usepackage{amsmath,epsfig}
\usepackage{stfloats}
\usepackage{breqn}
\usepackage{xcolor}
\usepackage{tabularx}
\usepackage{float}
\usepackage{url}
\usepackage{subcaption}
\usepackage{amsmath,amssymb,amsfonts}
\usepackage{graphicx}
\usepackage{booktabs}
\usepackage{caption}

\usepackage{multirow}
\usepackage{array}
\usepackage[margin=1in]{geometry}
\usepackage{array,tabularx}
\newcolumntype{Y}{>{\centering\arraybackslash}X}    
\newcolumntype{P}[1]{>{\raggedright\arraybackslash}p{#1}} 
\usepackage[style=ieee]{biblatex}

\newcommand{\R}{\mathbb{R}}

\definecolor{lightgray}{gray}{0.9}

\def\x{{\mathbf x}}
\def\y{{\mathbf y}}
\def\m{{\mathbf m}}
\def\p{{\mathbf p}}
\def\z{{\mathbf z}}
\def\a{{\mathbf a}}
\def\X{{\mathbf X}}
\def\Y{{\mathbf Y}}
\def\M{{\mathbf M}}
\def\P{{\mathbf P}}
\def\Q{{\mathbf Q}}
\def\W{{\mathbf W}}
\def\A{{\mathbf A}}
\def\E{{\mathbf E}}
\def\Rm{{\mathbf R}}
\def\Rmat{{\mathbf R}}
\def\V{{\mathbf V}}
\def\Z{{\mathbf Z}}

\def\U{{\mathcal U}}
\def\D{{\mathcal D}}

\title{SpectraMorph: Structured Latent Learning for\\Self-Supervised Hyperspectral Super-Resolution}

\author{
  Ritik Shah\\
  University of Massachusetts\\
  Amherst, MA 01003\\
  \texttt{rgshah@umass.edu}
  \and
  Marco F. Duarte\\
  University of Massachusetts\\
  Amherst, MA 01003\\
  \texttt{mduarte@umass.edu}
}

\begin{document}

\maketitle

\begin{abstract}
Hyperspectral sensors capture dense spectra per pixel but suffer from low spatial resolution, causing blurred boundaries and mixed-pixel effects. Co-registered companion sensors such as multispectral, RGB, or panchromatic cameras provide high-resolution spatial detail, motivating hyperspectral super-resolution through the fusion of hyperspectral and multispectral images (HSI-MSI). Existing deep learning based methods achieve strong performance but rely on opaque regressors that lack interpretability and often fail when the MSI has very few bands. We propose SpectraMorph, a physics-guided self-supervised fusion framework with a structured latent space. Instead of direct regression, SpectraMorph enforces an unmixing bottleneck: endmember signatures are extracted from the low-resolution HSI, and a compact multilayer perceptron predicts abundance-like maps from the MSI. Spectra are reconstructed by linear mixing, with training performed in a self-supervised manner via the MSI sensor’s spectral response function. SpectraMorph produces interpretable intermediates, trains in under a minute, and remains robust even with a single-band (pan-chromatic) MSI. Experiments on synthetic and real-world datasets show SpectraMorph consistently outperforming state-of-the-art unsupervised/self-supervised baselines while remaining very competitive against supervised baselines.
\end{abstract}

\section{Introduction}
\label{sec:intro}
Hyperspectral images (HSI) measure a dense reflectance spectrum at each pixel, enabling fine-grained material identification. In practice, however, optical and radiometric limits force a trade-off between spectral and spatial resolutions. Thus, the resulting HSI suffer from low spatial resolution: small structures blur together, mixed-pixel effects intensify, and downstream classification and mapping degrade. A widely adopted remedy is HSI–MSI fusion, also called hyperspectral super-resolution (HSR): given a low-resolution HSI (LR-HSI) and a co-registered high-resolution multispectral, RGB, or panchromatic image (HR-MSI/RGB/PAN) of the same scene, HSI-MSI fusion reconstructs a high-resolution HSI (HR-HSI).

Deep learning-based approaches to HSI-MSI fusion have improved reconstruction fidelity by learning direct image-to-image mappings. However, they often depend on scarce ground-truth HR-HSI and involve heavy architectures and extensive tuning. To mitigate supervision, physics-aware unsupervised schemes incorporate image formation knowledge -- point spread functions (PSF) which model blurring, spectral response functions (SRF), or mixing assumptions -- to constrain solutions. Yet PSF-based models lack robustness: the effective PSF is often unknown, variable, and difficult to estimate from data, leading to biased reconstructions and poor generalization outside controlled benchmarks. Further, most fusion methods combine HR-MSI with LR-HSI at the pixel level, implicitly assuming precise co-registration and a fixed blur model. In reality, HSI and MSI are often acquired at slightly different times (a phenomenon known as temporal misalignment), so objects move, shadows shift, illumination changes, and viewing geometry induces parallax, causing ghosting and halos in the super-resolved (SR) output.

To address these challenges, we recently proposed SpectraLift \cite{SpectraLift}, a per-pixel self-supervised framework for HSI–MSI fusion. A synthetic LR-MSI is generated by applying the MSI sensor’s SRF to the LR-HSI, and a compact multilayer perceptron (MLP) is trained to invert this SRF by mapping LR-MSI pixels back to LR-HSI pixels. In contrast to PSFs -- whose estimation is complicated by system optics, motion, and atmospheric effects -- SRFs are standardized, manufacturer-defined parameters that are routinely supplied for calibration purposes. The trained MLP is then applied to the HR-MSI to reconstruct an HR-HSI with fine spatial detail and rich spectra. SpectraLift eliminates the need for PSF estimation or HR-HSI ground truth, while remaining agnostic to blur, resolution, and temporal misalignment given that training inputs and targets originate from the same LR-HSI. While effective, SpectraLift faces two key limitations. First, when the MSI collapses to a single panchromatic band, the inversion becomes extremely ill-posed: an entire spectrum must be reconstructed from a single scalar intensity, often leading to severe artifacts. Second, although the final MSI-to-HSI mapping can be interpreted, the learning process itself is governed by a purely data-driven MLP, which is essentially a small black box with no interpretable intermediates.

We present SpectraMorph, which retains the self-supervised training regime of SpectraLift but replaces its opaque regressor with a structured, lightweight formulation. At its core, SpectraMorph uses a single hidden-layer MLP to predict abundance-like spatial maps from an MSI input, combined with a fixed endmember dictionary extracted from the LR-HSI via nonnegative matrix factorization (NMF). This unmixing bottleneck enforces physical meaning, yields interpretable intermediates, and achieves state-of-the-art HSR quality at a fraction of the computational cost of deep learning-based approaches. Beyond efficiency and interpretability, SpectraMorph also resolves the most ill-posed setting: when the HR input reduces to a single panchromatic band, SpectraMorph uses a coarse spectral prior (CSP) to provide stabilizing spectral context. Together, these advances establish SpectraMorph as a fusion framework that is accurate, interpretable, and computationally efficient. SpectraMorph's key contributions are:
\begin{itemize}
    \item \textbf{Structured unmixing bottleneck:} We replace opaque regression with a low-dimensional physics-guided latent space, using endmembers derived from LR-HSI via NMF and abundance-like maps to yield interpretable intermediates.
    \item \textbf{Coarse spectral prior:} We introduce a stabilizing prior that resolves the severely ill-posed  setting of the high resolution input containing just a single band (HR-PAN).
    \item \textbf{A lightweight, self-supervised machine learning model for HSR:} SpectraMorph retains the self-supervised regime of \cite{SpectraLift} while achieving state-of-the-art results on synthetic and real world benchmarks, with training times under one minute.
\end{itemize}
 
\section{Proposed Method}
\label{sec:method}

\begin{figure*}[!ht]
  \centering
  \begin{subfigure}[b]{0.95\textwidth}
    \centering
    \includegraphics[width=0.95\textwidth]{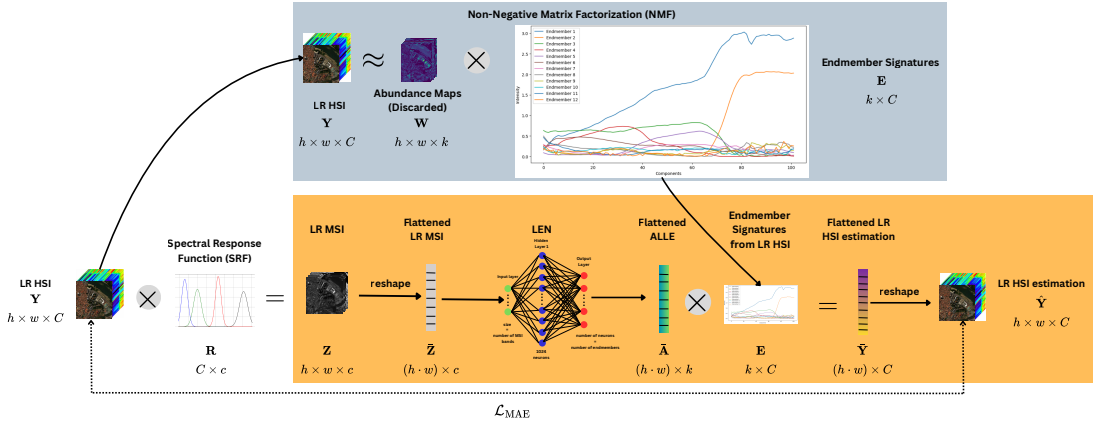}
    \label{fig:spectramoreph_train}
  \end{subfigure}

  \caption{{\small \sl SpectraMorph pipeline: the latent estimation network (LEN) produces a abundance-like latent estimate (ALLE) from a MSI pixel that is combined with a set of endmembers obtained from NMF (gray box) to estimate the corresponding HSI pixel. Training occurs at low resolution using a synthesized LR-MSI and the source LR-HSI. During inference, the same HSI pixel estimation process (orange box) is applied to the HR-MSI to obtain a HR-HSI estimate. Endmember dictionary is enlarged in the gray box.}} 
  \label{fig:spectramorph}
\end{figure*}

Let
$\Y  \in \mathbb{R}^{h \times w \times C}$
be the observed LR-HSI with \(C\) bands, and
$\M \in \mathbb{R}^{H \times W \times c}$ be
the HR-MSI with \(c < C\) bands. Under classical sensor models, these relate to the unknown HR-HSI
$\X \in \mathbb{R}^{H \times W \times C}$
via spatial and spectral degradations:
$\Y \approx \D_r\bigl(\mathcal{H}(\X, \Q)\bigr)$ and $\M \approx \X \times \Rmat$,
where \(\mathcal{H}(\X,\Q)\) denotes spatial convolution of the HSI image \(\X\) with the PSF \(\Q\), \(\D_r(\X)\) denotes spatial downsampling of the HSI \(\X\) by a factor \(r\), and \(\X \times \Rmat\) denotes the product of the HSI \(\X\) with the MSI SRF \(\Rmat\in\mathbb{R}^{C\times c}\) along the spectral (third) mode, i.e., for $i=1,\ldots,H$, $j = 1,\ldots,W$, $m=1,\ldots,c$, we have
\begin{align*}
    \M_{ijm}
    =
    \sum_{n=1}^{C} \X_{ijn}\,\Rmat_{nm}.
\end{align*}
In HSR, we estimate the unknown HR-HSI $\X$ from the paired observations ($\Y,\M$) under such degradation models.

\subsection{Leveraging NMF for Endmember Extraction}

We assume that pixel spectra are well approximated as a linear combination of a small set of endmember signatures,  
which is often obtained by applying NMF to the ``flattened'' HSI matrix $\bar{\Y} \in \R^{N\times C}$, $N=hw$, whose rows represent individual pixels:
\begin{equation}
\hat{\W},\hat{\E} = \operatorname*{argmin}_{\W\in\R^{N\times K},~{\E\in\R^{K\times C}},~(\E, \W)\ge 0}\ \|\bar{\Y}-\W\E\|_F^2,
\label{eq:nmf}
\end{equation}
for some unmixing complexity $K\ll C$. Here $\E$ denotes a matrix/dictionary containing the spectra for $K$ endmembers as its rows, representing spectra of identifiable material components, and $\W$ represents the weights/abundances of the endmembers in the $N$ image pixels. 
The optimization relies on a robust initialization (e.g., nonnegative double SVD~\cite{NNDSVD}) and sufficient iterations for stability. We preserve $\E$ to provide SpectraMorph with a fixed, interpretable spectral dictionary that anchors our unmixing bottleneck.

\subsection{Structured Latent Learning via Unmixing Bottleneck}
The orange region of Figure \ref{fig:spectramorph} represents the core learning process of SpectraMorph. We learn a lightweight MLP pixel-wise estimator $f_{\theta}$, dubbed the Latent Estimation Network (LEN), to map each MSI pixel to a $K$-dimensional abundance-like latent estimate (ALLE) representation:
$\a_n = f_{\Theta}(\z_n)\in\mathbb{R}^{K},$ for $n=1,\ldots,N,$
where the encoder inputs $\z_n$ are the pixels of the synthesized LR-MSI image \(\Z=\Y\times\Rmat\). The estimated spectra corresponding to the HSI pixels are obtained via linear mixing through the fixed dictionary $\E$, where the latent representation $\a_n$ provides the endmembers' weights: $\hat{\y}_n = \a_n\E$.

\begin{figure}[!t]
    \centering
    \includegraphics[width=\linewidth]{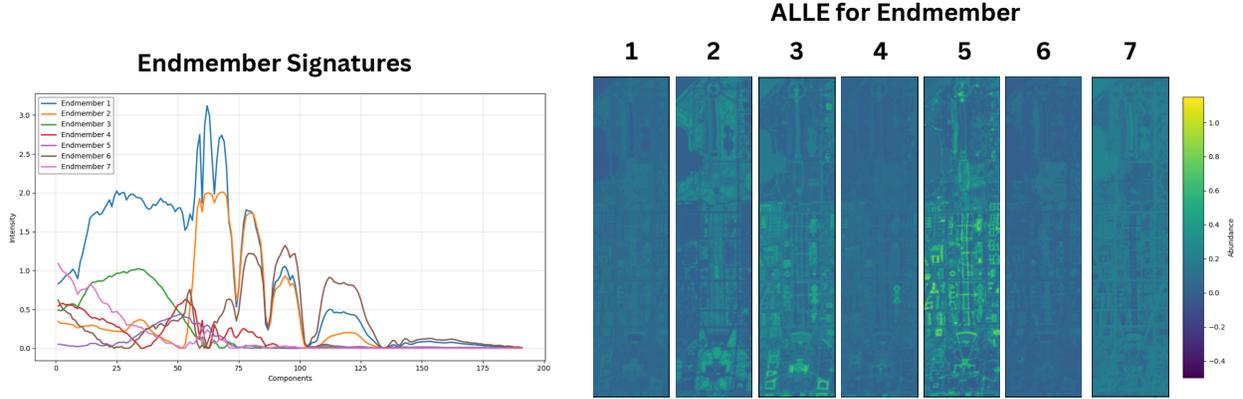} 
    \caption{\small Endmember Signatures $\E$ obtained from the LR-HSI and Abundance-like latent estimate (ALLE) maps $\A$ obtained from the HR-MSI during inference on Washington DC Mall.}
    \label{fig:abundances}
\end{figure}

In contrast to physical abundance maps, we \emph{do not} impose nonnegativity ($\a_n\ge 0$) nor sum-to-one ($\langle\a_n,\mathbf{1}\rangle=1$) constraints. The LEN's output is therefore \emph{abundance-like}: it is attached to an interpretable endmember dictionary $\E$ but is free to take any values that best explain the observed LR-HSI pixels under our training loss. This simplification relaxes modeling bias while preserving interpretability for the endmembers $\E$ and the obtained ``abundance-like'' maps $\A$ obtained by collecting the individual pixels $\a_n$ into a 3-D array, where the third dimension has length $k$ (cf.~Figure~\ref{fig:abundances}). The resulting ALLE should thus be viewed as latent activation coefficients rather than physically normalized material proportions. Their magnitudes and signs encode how strongly each endmember spectrum contributes to reconstructing a given pixel within the learned model. Values greater than one correspond to amplified activations, where an endmember’s spectral pattern is more pronounced than its nominal intensity in $\E$ -- often reflecting illumination variations or locally stronger material responses -- while negative values serve as compensatory terms that correct for correlations among endmembers or subtle nonlinear interactions not captured by a strictly additive model. In this formulation, the ALLEs provide a flexible and expressive latent representation that maintains spatial and spectral interpretability without the restrictive assumptions of classical abundance constraints.

We train the LEN using the synthesized LR-MSI $\Z$ and the LR-HSI $\Y$. 
The parameters $\Theta$ are optimized with a spectral mean-absolute-error (MAE) loss
\begin{equation}
\mathcal{L}(\Theta)= \frac{1}{N}\sum_{n=1}^{N}\left\|\,\hat{\y}_n - \y_n\,\right\|_{1}.
\label{eq:loss}
\end{equation}
This training process requires knowing only $\Y$ and $\Rm$ and avoids the need for the ground truth HR-HSI and PSF modeling; thus, learning is naturally tolerant to temporal misalignment and blur because the process is wholly formulated in the LR domain. That is, the synthesized LR-MSI $\Z$ will have the same blur as the LR-HSI $\Y$, and all objects as well as shadows will be in the same locations across the two images, enabling the LEN to be agnostic towards these and focus solely on estimating the ALLE~(cf.~Figure~\ref{fig:abundances}) that best explains the observations given the endmember dictionary $\E$.

To obtain the HR-HSI estimate $\hat{\X}$, we use the trained LEN on the pixels of the HR-MSI image $\M$: for $n = 1,\ldots, L$ (where $L=HW$), $\bar{\a}_n = f_{\Theta}(\m_n)$ and $\hat{\x}_n = \bar{\a}_n\E$. In summary, SpectraMorph replaces opaque regression with a structured unmixing bottleneck, where interpretability is enforced by the fixed endmember dictionary $\E$ and flexibility is provided by the learned ALLE. This structure allows the LEN to remain extremely compact, requiring only a single hidden layer and an output layer, while still achieving high-quality HSR. 

\subsection{Coarse Spectral Prior for Panchromatic Images}  
\label{sec:coarse-prior}  
\begin{figure}[!t]
    \centering
    \includegraphics[width=0.7\linewidth]{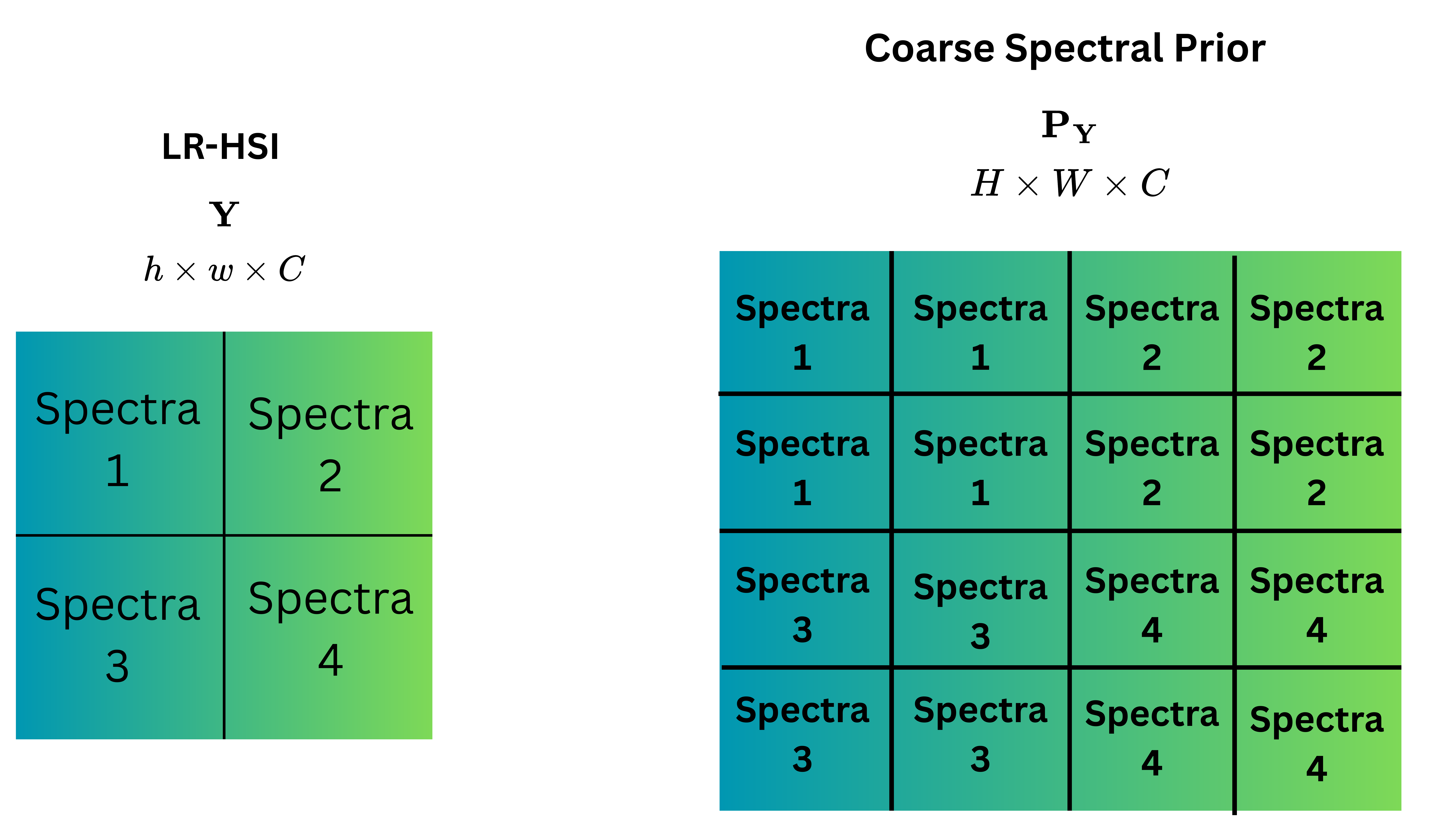} 
    \caption[Coarse spectral prior]{\small The coarse spectral prior replicates the spectra of the LR-HSI $\Y$ to obtain a CSP $\P_\Y$ that serves as side information for inference from a panchromatic image. Here, $\Y \approx \D_2\bigl(\mathcal{H}(\X, \Q)\bigr)$ ($r=2$), hence the spectra of each pixel of $\Y$ is replicated $2\times2$ times to obtain $\P_Y$.}
    \label{fig:coarse_prior}
\end{figure}

For the HR-PAN case (\(c=1\)), spectral inversion is extremely ill-posed. We stabilize training by injecting a \emph{coarse spectral prior} (CSP) from the LR-HSI as side information.  During training, we spatially downsample the LR-HSI by a downsampling factor \((s=2,4,8,16,\ldots)\), which we denote by $\V = \D_s(\Y)$ of shape $(h/s, w/s, C)$, and replicate each of its pixels \(s \times s\) times (e.g., $s$ times in each dimension), which we denote by  
$\P_\V = \U_s(\V)$.
We refer to the resulting $h\times w\times C$ image $\P_\V$ as the CSP of $\V$. Thus, each pixel block in \(\P_\V\) carries the same single spectrum, and the CSP retains only low-frequency spatial context. The LEN is then modified to have a concatenation of a LR-MSI pixel $\z_n$ and the corresponding CSP pixel ${\overline{\p_\V}}_n$ as its input, and the rest of the SpectraMorph pipeline remains unchanged.
At inference, we compute the LR-HSI's CSP as $\P_\Y = \U_r(\Y)$, illustrated in Figure~\ref{fig:coarse_prior}, where $r = H/h=W/w$ is the HR-to-LR downsampling factor so that $\P_\Y$ has the same spatial resolution as $\M$. We then feed the concatenation of $\M$ and $\P_\Y$ to the modified LEN to estimate $\hat{\X}$.
 
\subsection{Interpretability, Assumptions, and Limitations}
\textbf{Interpretability.}
A notable advantage of the proposed formulation is its inherent interpretability. The rows of $\E$ can be directly visualized as endmember spectra~(cf.~Figure~\ref{fig:abundances}), each corresponding to physically meaningful materials such as vegetation, soil, or man-made surfaces like roofing. These spectra serve as an intuitive “dictionary” that links the model to the underlying physics of the scene. Complementing this, the latent maps (i.e., the channels of $\a$) act as abundance-like spatial fields, describing how strongly each endmember contributes to explaining the observations at every pixel. Unlike traditional spectral unmixing approaches, our framework does not explicitly impose nonnegativity or sum-to-one constraints on these maps. Nevertheless, the resulting representations remain interpretable, since the latent activations naturally highlight the presence and spatial distribution of different materials across the scene. This dual representation -- spectrally in $\E$ and spatially in $\A$ -- provides a transparent and physically grounded intermediate layer, while still allowing for a compact and flexible model design. As a result, the fusion process not only produces high-resolution reconstructions but also offers human-readable insights into how those reconstructions are formed.

\noindent\textbf{Assumptions.}
Our framework relies on a few standard but practically reasonable assumptions:

($i$) \textit{Spectral response function:} We assume that the SRF $\R$ of the multispectral sensor is either exactly known or can be accurately approximated. In most cases, this information is well-documented by the sensor manufacturer and is routinely employed during instrument calibration. Even when the exact SRF is unavailable, common practice shows that Gaussian approximations of widely used satellite SRFs provide sufficiently accurate results, and thus do not significantly hinder the performance of the proposed method.

($ii$) \textit{Number of endmembers:} We assume that a reasonable estimate of the number of endmembers $K$ is available. This can be obtained from prior domain knowledge, exploratory analysis, or standard model selection criteria. In our experiments, we adopt a pragmatic approach by setting $K$ equal to the number of annotated land-cover classes in the benchmark datasets. For instance, in the Pavia Center dataset, which contains nine classes, we fix $K=9$. This choice ensures a direct alignment between the learned spectral bases and the semantic categories of interest, while still being consistent with typical hyperspectral unmixing practice.

($iii$) \textit{Data preprocessing:} We assume that the hyperspectral data $\Y$ and the multispectral data $\M$ have undergone standard radiometric corrections and preprocessing steps, so that they are directly compatible in terms of scale and physical units. This is a common requirement in hyperspectral–multispectral fusion pipelines, as preprocessing mitigates sensor-specific artifacts and ensures that the two modalities can be meaningfully integrated within a unified model.

\noindent\textbf{Limitations:}
While the proposed framework offers strong interpretability and competitive performance, it is important to acknowledge its practical limitations:

($i$) \textit{Choice of endmember count:} The number of endmembers $K$ must be selected with care. Too few endmembers may lead to underfitting, where important spectral variability is not captured, while too many may cause overfitting or introduce spurious bases that reduce interpretability. Although domain knowledge and class labels provide guidance, the tuning of $K$ remains a sensitive step in practice.

($ii$) \textit{Coverage of materials:} Because the endmember extraction is performed on the low-resolution hyperspectral image $\Y$, any materials that are present in pure form only in the high-resolution image $\X$ but absent or severely mixed in $\Y$ may not be represented in the learned endmember matrix $\E$. This limitation can result in missing or biased reconstructions when novel materials appear at finer spatial scales.

($iii$) \textit{Dependence on SRF accuracy:} The method assumes that the spectral response function $\R$ used to generate the synthetic low-resolution multispectral observation $\Z$ closely matches the true MSI sensor characteristics. Substantial mismatches between the assumed and actual SRF can lead to biased supervision signals, degrading both reconstruction fidelity and interpretability.

($iv$) \textit{Scene-specific training:} Since the endmember matrix $\E$ is extracted in a scene-dependent manner, the model must be retrained for each new LR-HSI/HR-MSI pair. A network trained on one scene cannot be directly transferred to another without re-extracting the endmembers, limiting the potential for cross-scene generalization which is a common limitation among state-of-the-art unsupervised/self-supervised HSR methods. 

\section{Experiments and Analysis}
\label{ssec:experiments}
We evaluate SpectraMorph on synthetic and real-world benchmarks. We compare against nine state-of-the-art baselines: five unsupervised (SSSR~\cite{SSSR}, MIAE~\cite{MIAE}, C2FF~\cite{C2FF}, SDP~\cite{SDP}, SpectraLift~\cite{SpectraLift}) and four supervised (GuidedNet~\cite{GuidedNet}, FeINFN~\cite{FeINFN}, FusFormer~\cite{FusFormer}, MIMO-SST~\cite{MIMO}).   
All datasets, precomputed results, and detailed instructions for replication are available at \url{https://github.com/ritikgshah/SpectraMorph}. We have made every effort to ensure reproducibility by providing pre-executed Jupyter notebooks that were used to obtain the results showcased in this paper, Python scripts for end-to-end runs, configuration files, and environment replication scripts for seamless setup. This enables reviewers and practitioners to verify results and explore extensions with minimal friction.

\subsection{SpectraMorph Implementation Details}
SpectraMorph is implemented using the TensorFlow framework and trained with the Adam optimizer. For learning rate scheduling, we use the One-Cycle Learning Rate policy, as it accelerates convergence and promotes stability. We tuned the scheduler parameters -- specifically, the initial, maximum, and final learning rates. The exact learning rate configurations used in each of our experiments are provided in the pre-executed Jupyter notebooks, which are available in our GitHub repository. 

\subsection{Quality Metrics} For synthetic data, we evaluate SpectraMorph using six widely adopted metrics that capture complementary aspects of reconstruction quality in HSR:

\begin{itemize}
    \item {\bf Root Mean Squared Error (RMSE):} Measures the average pixel-wise difference between the reconstructed hyperspectral image \(\hat{\X}\) and the ground truth \(\X\). Lower RMSE indicates higher reconstruction fidelity.
    \[
    \mathrm{RMSE}(\X,\hat{\X}) = \sqrt{\frac{1}{HWC}\sum_{i=1}^{H}\sum_{j=1}^{W}\sum_{k=1}^{C} \bigl(\hat{\X}_{ijk} - \X_{ijk}\bigr)^2}.
    \]

    \item {\bf Peak Signal-to-Noise Ratio (PSNR):} Quantifies the ratio between the maximum possible pixel value and the power of the reconstruction error, expressed in decibels (dB). Higher PSNR indicates better perceptual quality.
    \[
    \mathrm{PSNR}(\X,\hat{\X}) = 10 \log_{10}\left(\frac{\mathrm{MAX}^2}{\mathrm{RMSE}(\X,\hat{\X})^2}\right),
    \]
    where \(\mathrm{MAX}\) is the maximum possible pixel value (e.g., \(2^{16}-1\) for 16-bit images).

    \item {\bf Structural Similarity Index Measure (SSIM):} Assesses perceptual similarity by comparing luminance, contrast, and structure between \(\hat{\X}\) and \(\X\). Values close to 1 indicate high structural similarity.
    \[
    \mathrm{SSIM}(\X, \hat{\X}) = \frac{(2\mu_X\mu_{\hat{X}} + c_1)(2\sigma_{X\hat{X}} + c_2)}{(\mu_X^2 + \mu_{\hat{X}}^2 + c_1)(\sigma_X^2 + \sigma_{\hat{X}}^2 + c_2)},
    \]
    where \(\mu\), \(\sigma^2\), and \(\sigma_{X\hat{X}}\) are means, variances, and covariances, and \(c_1, c_2\) are small constants to stabilize the denominator.

    \item {\bf Universal Image Quality Index (UIQI):} Measures similarity in terms of luminance, contrast, and structure. Values range from \(-1\) to 1, with higher values indicating better quality.
    \[
    \mathrm{UIQI}(\X,\hat{\X}) = \frac{4\sigma_{X\hat{X}}\mu_X\mu_{\hat{X}}}{\left(\sigma_X^2 + \sigma_{\hat{X}}^2\right)\left(\mu_X^2 + \mu_{\hat{X}}^2\right)}.
    \]

    \item {\bf Erreur Relative Globale Adimensionnelle de Synth\`{e}se (ERGAS):} Provides a global indication of the relative error, normalized by the mean reflectance, and is commonly used in remote sensing. Lower ERGAS indicates higher reconstruction accuracy.
    \[
    \mathrm{ERGAS}(\X,\hat{\X}) = 100~\frac{r}{s}\sqrt{\frac{1}{C}\sum_{k=1}^{C}\frac{\mathrm{RMSE}_k(\X, \hat{\X})^2}{\mu_k^2}},
    \]
    where \(r/s\) is the ratio of spatial resolutions between \(\hat{\X}\) and \(\X\), and \(\mathrm{RMSE}_k(\X, \hat{\X})\) and \(\mu_k\) are the RMSE and mean of band \(k\), respectively.

    \item {\bf Spectral Angle Mapper (SAM):} Computes the mean spectral angle (in degrees) between estimated and ground-truth spectral vectors at each pixel. It measures spectral similarity, with smaller angles indicating better fidelity. To ensure numerical stability, the arccos argument is clipped to avoid numerical issues.

    \[
    \mathrm{SAM}(\X, \hat{\X}) = \frac{1}{HW} \sum_{i=1}^{H}\sum_{j=1}^{W} \left( \frac{180}{\pi} \arccos\left( \min\left( \frac{\langle \X_{ij:}, \hat{\X}_{ij:} \rangle}{\|\X_{ij:}\|_2 \|\hat{\X}_{ij:}\|_2 + \epsilon}, 1 - \delta \right) \right) \right),
    \]
    
    where $\X_{ij:}$ and $\hat{\X}_{ij:}$ are the spectral vectors at pixel $(i,j)$, and $\epsilon$ and $\delta$ are small constants to avoid division by zero and numerical overflow, respectively (e.g., $\epsilon=10^{-8}$, $\delta=10^{-9}$).

\end{itemize}

We also profile execution time, number of parameters, inference peak GPU memory used and inference FLOPs to assess model complexity. For the inference GPU memory used and inference FLOPs, we consider these for a single forward pass of the inputs through the model. In the tables below, we specify whether the best value of a metric is lower ($\downarrow$) or higher ($\uparrow$).

\subsection{Synthetic Datasets}
\begin{figure*}[!t]
    \centering
    \includegraphics[width=\textwidth]{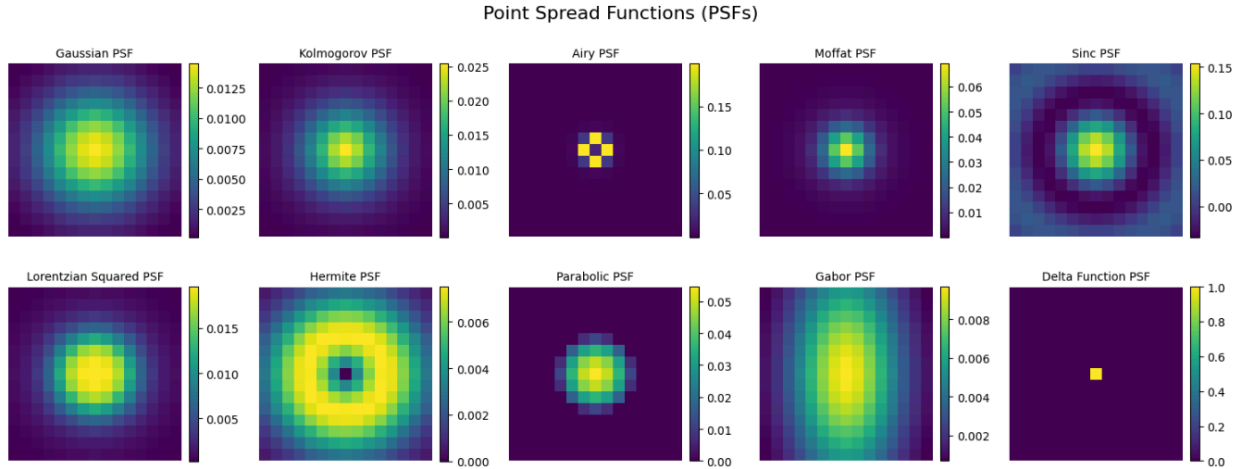} 
    \caption{Point Spread Functions used for Synthetic LR HSI generation}
    \label{fig:psf}
\end{figure*}

Following Wald's protocol~\cite{wald}, for each ground-truth HSI (GT): Washington DC Mall (DC), Kennedy Space Center (KSC), Botswana, Pavia University (Pavia U), and Pavia Center (Pavia), we generate LR-HSI by first applying spatial convolution of the GT with one of ten PSFs shown in Figure \ref{fig:psf} (Gaussian, Kolmogorov, Airy, Moffat, Sinc, Lorentzian Squared, Hermite, Parabolic, Gabor, Delta), all with kernel size of (15,15). We then downsample the convolution output by \(r\in\{4,8,16,32\}\), and finally add Gaussian white noise with SNR matched to \(r\): (4, 35 dB), (8, 30 dB), (16, 25 dB), (32, 20 dB), yielding \(10\times4=40\) LR-HSIs for each GT. HR-MSI are synthesized by applying SRFs for \(c\in\{1,3,4\}\) (IKONOS), \(c=8\) (WorldView-2), and \(c=16\) (WorldView-3), then adding Gaussian white noise with SNR=40 dB, producing 5 HR-MSIs for each GT. 

We consider 80 LR-HSI/HR-MSI pairs per GT: 10 PSFs $\times$ 8 representative $(r,c)$ configurations \{(4,4), (8,4), (16,4), (32,4), (8,1), (8,3), (8,8), (8,16)\}. We apply all the degradations to the GT normalized between [0,1].
Supervised methods are trained on 75\% crops and tested on 25\% crops for each LR-HSI-HR-MSI input pair. Spatial dimensions for the training and testing crops are constrained to be  multiples of 32, ensuring compatibility with models that require integer downsampling without residual pixels. Unsupervised methods use full images but report metrics on the same 25\% test regions. To support a more fair comparison across supervised and unsupervised approaches, the latter are also given access to the PSF and SRF used for LR-HSI and HR-MSI generation.

\subsection{Results} Tables \ref{tab:dc_quality}-\ref{tab:botswana_quality} report mean quality and complexity metrics over these 80 LR-HSI/HR-MSI pairs for the Washington DC Mall, Kennedy Space Center, Pavia University, Pavia Center, and Botswana benchmarks, respectively. Metrics for the supervised methods have been shaded in gray. Across all five synthetic datasets, SpectraMorph outperforms unsupervised baselines on almost all quality measures. In some cases, our method trades slightly worse metrics, compared to SpectraLift, for substantial gains in interpretability and efficiency, confirming the metric–interpretability tradeoff. Supervised methods enjoyed a modest advantage -- they were trained on a 75\% crop of the same scene (distinct from the 25\% test region), allowing them to learn scene-specific spectral behaviors. Despite lacking this advantage, SpectraMorph surpasses most supervised approaches. Only MIMO-SST consistently edges ahead, but with a more complex method relying on ``black box'' models with knowledge of the scene specific spectra. Notably, SpectraMorph achieves the lowest SAM on almost all of our synthetic datasets, underscoring its ability to deliver both high spatial detail and faithful spectral reconstructions without supervision or PSF knowledge. 

SpectraMorph is the only method to combine strong accuracy with extreme efficiency, training in under one minute per scene, whereas other baselines range from hundreds to tens of thousands of seconds. SpectraMorph's inference cost in FLOPs and GPU memory required during inference is close to those of the most lightweight approaches and significantly lower than that of more complex methods. SpectraMorph's computational complexity is even lower than that of SpectraLift's in all scenarios except when $(r, c) = (8, 1)$ owing to the use of the CSP. Table \ref{tab:csp_ablation} shows a considerable increase in the number of parameters and FLOPs when the CSP is used compared to the study that omits employing the CSP. This lightweight design with high super resolution quality metrics makes SpectraMorph uniquely suitable for practical, near real-time deployment on real-world systems. 

\begin{table*}[t]
  \centering
  \setlength\tabcolsep{4pt}          
  \renewcommand{\arraystretch}{1.1}  
  \captionsetup{font=small, justification=centering}
  \caption[Quality measures]{Quality measures for the Washington DC Mall dataset: Mean value of 80 LR-HSI/HR-MSI configurations (\textbf{best in bold}, \textit{second best in italics}; supervised methods are shaded).}
  \label{tab:dc_quality}
  
  \resizebox{\columnwidth}{!}{%
  \begin{tabular}{|l|c|c|c|c|c|c|c|c|c|c|}
    \hline
    \textbf{Method} & \textbf{RMSE $\downarrow$} & \textbf{PSNR $\uparrow$} & \textbf{SSIM $\uparrow$} 
    & \textbf{UIQI $\uparrow$} & \textbf{ERGAS $\downarrow$} & \textbf{SAM $\downarrow$} 
    & \textbf{Time (s) $\downarrow$} & \textbf{Params (M) $\downarrow$} & \textbf{FLOPs (G) $\downarrow$} 
    & \textbf{GPU Mem (MB) $\downarrow$} \\
    \hline
    MIAE         & 0.03611 & 30.28 & 0.934 & 0.968 & 5.33  & 5.21  & 212.69  & \textbf{0.0218} & 8.46      & 707.54 \\
    \hline
    C2FF         & 0.02681 & 34.74 & 0.960 & 0.975 & 4.73  & 3.53  & 78.54   & 0.0979 & \textit{7.29}    & 948.88 \\
    \hline
    SDP          & 0.03029 & 31.99 & 0.939 & 0.907 & 16.59 & 4.29  & 799.75  & 6.4546 & 511.52    & 5370.37 \\
    \hline
    SSSR         & 0.05024 & 27.98 & 0.908 & 0.938 & 7.49  & 5.83  & \textit{67.24}   & \textit{0.0334} & \textbf{0.000067} & \textit{286.31} \\
    \hline
    SpectraLift  & 0.02344 & \textbf{35.96} & 0.967 & 0.969 & 5.87  & \textit{3.22}  & 91.20   & 0.0336 & 26.28     & \textit{286.31} \\
    \hline
    \rowcolor{lightgray}
    GuidedNet    & 0.03494 & 29.98 & 0.919 & 0.872 & 30.30 & 4.81  & 398.07  & 6.7111 & 178.41    & \textbf{81.52} \\
    \hline
    \rowcolor{lightgray}
    FeINFN       & 0.02558 & 32.66 & 0.966 & 0.956 & 7.83  & 3.79  & 3018.80 & 3.7021 & 251.70    & 982.47 \\
    \hline
    \rowcolor{lightgray}
    FusFormer    & 0.02555 & 32.55 & 0.876 & 0.688 & 17.34 & 3.89  & 11110.07 & 0.1883 & 946.28    & 6328.36 \\
    \hline
    \rowcolor{lightgray}
    MIMO-SST         & \textit{0.02193} & 35.21 & \textit{0.969} & \textit{0.982} & \textbf{2.92}  & 3.29  & 376.34  & 2.1879 & 56.37     & 393.74 \\
    \hline
    SpectraMorph & \textbf{0.01943} & \textit{35.74} & \textbf{0.976} & \textbf{0.984} & \textit{3.38} & \textbf{2.63} & \textbf{42.96} & 0.0383 & 30.73 & 300.97 \\
    \hline
    \hline
    Oracle & 0.01778 & 36.44 & 0.979 & 0.985 & 3.32 & 2.21 & - & - & - & - \\
    \hline
  \end{tabular}}
\end{table*}

\begin{table*}[t]
  \centering
  \footnotesize
  \setlength\tabcolsep{4pt}
  \renewcommand{\arraystretch}{1.1}
  \captionsetup{font=small, justification=centering}
  \caption{Quality measures for the Kennedy Space Center dataset: Mean value of 80 LR HSI-HR MSI configurations (\textbf{best in bold}, \textit{second best in italics}; supervised methods are shaded).}
  \label{tab:ksc_quality}
   \resizebox{\columnwidth}{!}{%
   \begin{tabular}{|l|c|c|c|c|c|c|c|c|c|c|}
    \hline
    \textbf{Method} & \textbf{RMSE $\downarrow$} & \textbf{PSNR $\uparrow$} & \textbf{SSIM $\uparrow$}
    & \textbf{UIQI $\uparrow$} & \textbf{ERGAS $\downarrow$} & \textbf{SAM $\downarrow$}
    & \textbf{Time (s) $\downarrow$} & \textbf{Params (M) $\downarrow$} & \textbf{FLOPs (G) $\downarrow$}
    & \textbf{GPU Mem (MB) $\downarrow$} \\
    \hline
    MIAE         & 0.04855 & 26.41 & 0.907 & \textbf{0.962} & 8.54 & 8.88 & 189.24 & 0.0996 & 31.12    & 711.06 \\
    \hline
    C2FF         & 0.04540 & 26.97 & 0.922 & 0.956          & 8.50 & 8.90 & 77.24  & 0.0913 & \textit{5.42}   & 707.95 \\
    \hline
    SDP          & 0.04659 & 26.73 & 0.906 & 0.949          & 8.63 & 9.65 & 643.65 & 6.1322 & 388.55    & 4209.39 \\
    \hline
    SSSR         & 0.05767 & 25.05 & 0.851 & 0.930          & 9.54 & 10.08& \textit{70.70} & \textbf{0.0284} & \textbf{0.000057} & \textit{211.06} \\
    \hline
    SpectraLift  & \textit{0.04472} & \textit{27.10} & \textit{0.925} & 0.957 & \textit{8.41} & \textit{8.70} & 92.24 & \textit{0.0327} & 20.42 & \textit{211.06} \\
    \hline
    \rowcolor{lightgray}
    GuidedNet    & 0.04951 & 26.11 & 0.913 & 0.956          & 9.17 & 9.25 & 374.44 & 6.0301 & 176.43    & \textbf{71.32} \\
    \hline
    \rowcolor{lightgray}
    FeINFN       & 0.05013 & 26.02 & 0.902 & 0.932          & 9.07 & 10.77& 2206.34& 3.6520 & 261.73    & 984.15 \\
    \hline
    \rowcolor{lightgray}
    FusFormer    & 0.05124 & 25.82 & 0.888 & 0.921          & 9.26 & 11.57& 6976.85& 0.1811 & 787.98    & 6322.09 \\
    \hline
    \rowcolor{lightgray}
    MIMO-SST         & 0.04660 & 26.68 & 0.917 & 0.941          & 8.70 & 9.47 & 305.21 & 2.1361 & 56.79     & 363.68 \\
    \hline
    SpectraMorph & \textbf{0.04299} & \textbf{27.37} & \textbf{0.934} & \textit{0.961} & \textbf{8.24} & \textbf{8.22} & \textbf{42.31} & 0.0425 & 27.84 & 223.82 \\
    \hline
    \hline
    Oracle         & 0.04626 & 26.71 & 0.941 & 0.963 & 8.71 & 8.59 & - & - & - & - \\
    \hline
  \end{tabular}}
\end{table*}

\begin{table*}[t]
  \centering
  \footnotesize
  \setlength\tabcolsep{4pt}
  \renewcommand{\arraystretch}{1.1}
  \captionsetup{font=small, justification=centering}
  \caption{Quality measures for the Pavia University dataset: Mean value of 80 LR HSI-HR MSI configurations (\textbf{best in bold}, \textit{second best in italics}; supervised methods are shaded).}
  \label{tab:pavia_u_quality}
  \resizebox{\columnwidth}{!}{%
  \begin{tabular}{|l|c|c|c|c|c|c|c|c|c|c|}
    \hline
    \textbf{Method} & \textbf{RMSE $\downarrow$} & \textbf{PSNR $\uparrow$} & \textbf{SSIM $\uparrow$}
    & \textbf{UIQI $\uparrow$} & \textbf{ERGAS $\downarrow$} & \textbf{SAM $\downarrow$}
    & \textbf{Time (s) $\downarrow$} & \textbf{Params (M) $\downarrow$} & \textbf{FLOPs (G) $\downarrow$}
    & \textbf{GPU Mem (MB) $\downarrow$} \\
    \hline
    MIAE         & 0.03775 & 29.62 & 0.906 & 0.984 & 3.64 & 5.01 & 99.99  & 0.0880 & 18.11   & 445.91 \\
    \hline
    C2FF         & 0.03618 & 31.70 & 0.914 & 0.976 & 3.36 & 4.63 & 73.24  & 0.0590 & \textit{2.25}   & 293.47 \\
    \hline
    SDP          & 0.03088 & 31.88 & 0.917 & 0.982 & 3.14 & 4.29 & 385.58 & 4.6211 & 192.66  & 2491.03 \\
    \hline
    SSSR         & 0.05851 & 26.31 & 0.829 & 0.950 & 5.46 & 5.68 & \textit{66.48} & \textbf{0.0099} & \textbf{0.000020} & \textit{81.49} \\
    \hline
    SpectraLift  & 0.03073 & 32.79 & 0.928 & 0.980 & 2.99 & 4.18 & 91.03  & \textit{0.0279} & 11.52  & \textit{81.49} \\
    \hline
    \rowcolor{lightgray}
    GuidedNet    & 0.03060 & 30.80 & 0.920 & 0.986 & 3.36 & 4.19 & 243.72 & 3.4673 & 68.62   & \textbf{22.63} \\
    \hline
    \rowcolor{lightgray}
    FeINFN       & 0.02789 & 31.82 & 0.933 & 0.982 & 3.04 & 4.18 & 1187.35& 3.4082 & 127.63  & 381.94 \\
    \hline
    \rowcolor{lightgray}
    FusFormer    & 0.03017 & 31.00 & 0.929 & 0.983 & 3.32 & 4.22 & 5556.52& 0.1460 & 471.06  & 6231.11 \\
    \hline
    \rowcolor{lightgray}
    MIMO-SST         & \textbf{0.02236} & \textbf{34.34} & \textbf{0.945} & \textbf{0.989} & \textbf{2.49} & \textbf{3.33} & 221.63 & 1.8836 & 22.93  & 153.75 \\
    \hline
    SpectraMorph & \textit{0.02553} & \textit{33.33} & \textit{0.939} & \textit{0.987} & \textit{2.63} & \textit{3.46} & \textbf{40.45} & 0.0291 & 12.23 & 89.19 \\
    \hline
    \hline
    Oracle       & 0.02157 & 34.70 & 0.948 & 0.992 & 2.37 & 2.95 & - & - & - & - \\
    \hline
  \end{tabular}}
\end{table*}

\begin{table*}[t]
  \centering
  \footnotesize
  \setlength\tabcolsep{4pt}
  \renewcommand{\arraystretch}{1.1}
  \captionsetup{font=small, justification=centering}
  \caption{Quality measures for the Pavia Center dataset: Mean value of 80 LR HSI-HR MSI configurations (\textbf{best in bold}, \textit{second best in italics}; supervised methods are shaded).}
  \label{tab:pavia_c_quality}
  \resizebox{\columnwidth}{!}{%
  \begin{tabular}{|l|c|c|c|c|c|c|c|c|c|c|}
    \hline
    \textbf{Method} & \textbf{RMSE $\downarrow$} & \textbf{PSNR $\uparrow$} & \textbf{SSIM $\uparrow$}
    & \textbf{UIQI $\uparrow$} & \textbf{ERGAS $\downarrow$} & \textbf{SAM $\downarrow$}
    & \textbf{Time (s) $\downarrow$} & \textbf{Params (M) $\downarrow$} & \textbf{FLOPs (G) $\downarrow$}
    & \textbf{GPU Mem (MB) $\downarrow$} \\
    \hline
    MIAE         & 0.04344 & 28.67 & 0.901 & 0.984 & 3.58 & 6.99 & 314.86 & 0.0878 & 68.30    & 1677.58 \\
    \hline
    C2FF         & 0.02861 & 33.38 & 0.956 & 0.988 & 2.46 & 4.90 & 85.77  & 0.0586 & \textit{8.43}    & 1094.10 \\
    \hline
    SDP          & 0.02979 & 32.11 & 0.952 & 0.988 & 2.61 & 5.58 & 1085.04& 4.6011 & 724.77   & 9375.14 \\
    \hline
    SSSR         & 0.06399 & 25.67 & 0.839 & 0.958 & 5.26 & 6.28 & \textit{67.55} & \textbf{0.0098} & \textbf{0.000020} & \textit{304.91} \\
    \hline
    SpectraLift  & 0.02716 & 33.88 & 0.958 & 0.988 & 2.36 & 4.80 & 90.84  & \textit{0.0278} & 43.41 & \textit{304.91} \\
    \hline
    \rowcolor{lightgray}
    GuidedNet    & 0.02929 & 31.10 & 0.954 & 0.992 & 2.68 & 4.98 & 723.75 & 3.4408 & 263.18   & \textbf{92.44} \\
    \hline
    \rowcolor{lightgray}
    FeINFN       & 0.02532 & 32.48 & 0.964 & 0.992 & 2.33 & 4.83 & 5131.20& 3.4048 & 490.85   & 1459.95 \\
    \hline
    \rowcolor{lightgray}
    FusFormer    & 0.02262 & 33.58 & \textbf{0.968} & \textit{0.993} & 2.09 & 4.49 & 19432.89& 0.1455 & 1884.16 & 6321.71 \\
    \hline
    \rowcolor{lightgray}
    MIMO-SST         & \textit{0.02168} & \textbf{35.02} & 0.963 & \textit{0.993} & \textit{2.01} & \textit{4.19} & 706.33 & 1.8802 & 87.91   & 575.90 \\
    \hline
    SpectraMorph & \textbf{0.02163} & \textit{34.42} & \textit{0.967} & \textbf{0.995} & \textbf{2.00} & \textbf{3.83} & \textbf{42.08} & 0.0289 & 45.98 & 331.82 \\
    \hline
    \hline
    Oracle       & 0.01997 & 35.18 & 0.969 & 0.995 & 1.89 & 3.52 & - & - & - & - \\
    \hline
  \end{tabular}}
\end{table*}

\begin{table*}[t]
  \centering
  \footnotesize
  \setlength\tabcolsep{4pt}
  \renewcommand{\arraystretch}{1.1}
  \captionsetup{font=small, justification=centering}
  \caption{Quality measures for the Botswana dataset: Mean value of 80 LR HSI-HR MSI configurations (\textbf{best in bold}, \textit{second best in italics}; supervised methods are shaded).}
  \label{tab:botswana_quality}
  \resizebox{\columnwidth}{!}{%
  \begin{tabular}{|l|c|c|c|c|c|c|c|c|c|c|}
    \hline
    \textbf{Method} & \textbf{RMSE $\downarrow$} & \textbf{PSNR $\uparrow$} & \textbf{SSIM $\uparrow$}
    & \textbf{UIQI $\uparrow$} & \textbf{ERGAS $\downarrow$} & \textbf{SAM $\downarrow$}
    & \textbf{Time (s) $\downarrow$} & \textbf{Params (M) $\downarrow$} & \textbf{FLOPs (G) $\downarrow$}
    & \textbf{GPU Mem (MB) $\downarrow$} \\
    \hline
    MIAE         & 0.01838 & 35.06 & 0.954 & \textit{0.997} & 1.75 & 1.77 & 162.78 & \textbf{0.0190} & 7.09    & 552.00 \\
    \hline
    C2FF         & 0.01572 & 36.51 & 0.970 & \textbf{0.998} & 1.48 & 1.59 & \textit{74.82} & 0.0776 & \textit{5.48}   & 716.48 \\
    \hline
    SDP          & 0.01863 & 34.82 & 0.951 & 0.995          & 2.29 & 1.97 & 698.01  & 5.4788 & 416.77  & 4835.41 \\
    \hline
    SSSR         & 0.02050 & 34.20 & 0.953 & 0.996          & 1.93 & 1.69 & 86.12   & \textit{0.0194} & \textbf{0.000039} & \textit{208.97} \\
    \hline
    SpectraLift  & 0.01452 & 37.04 & \textbf{0.974} & \textit{0.997} & 1.60 & \textit{1.46} & 93.97  & 0.0306 & 23.03  & 209.00 \\
    \hline
    \rowcolor{lightgray}
    GuidedNet    & 0.01970 & 34.25 & 0.927 & 0.986          & 4.11 & 1.93 & 439.14  & 4.7894 & 162.62  & \textbf{68.65} \\
    \hline
    \rowcolor{lightgray}
    FeINFN       & 0.02250 & 33.44 & 0.946 & 0.995          & 1.95 & 2.53 & 2718.46 & 3.5485 & 266.91  & 919.15 \\
    \hline
    \rowcolor{lightgray}
    FusFormer    & 0.02937 & 31.10 & 0.780 & 0.815          & 11.23& 3.35 & 8333.27 & 0.1662 & 944.11  & 6299.68 \\
    \hline
    \rowcolor{lightgray}
    MIMO-SST         & \textbf{0.01346} & \textbf{37.76} & \textit{0.973} & \textbf{0.998} & \textbf{1.35} & 1.54 & 365.49 & 2.0289 & 53.87  & 367.98 \\
    \hline
    SpectraMorph & \textit{0.01423} & \textit{37.14} & \textbf{0.974} & \textbf{0.998} & \textit{1.41} & \textbf{1.44} & \textbf{41.43} & 0.0396 & 31.04 & 229.18 \\
    \hline
    \hline
    Oracle          & 0.01178 & 38.87 & 0.977 & 0.998 & 1.27 & 1.29 & - & - & - & - \\
    \hline
  \end{tabular}}
\end{table*}

\begin{figure*}[!ht]
  \centering

  \begin{subfigure}{\textwidth}
    \includegraphics[width=\textwidth]{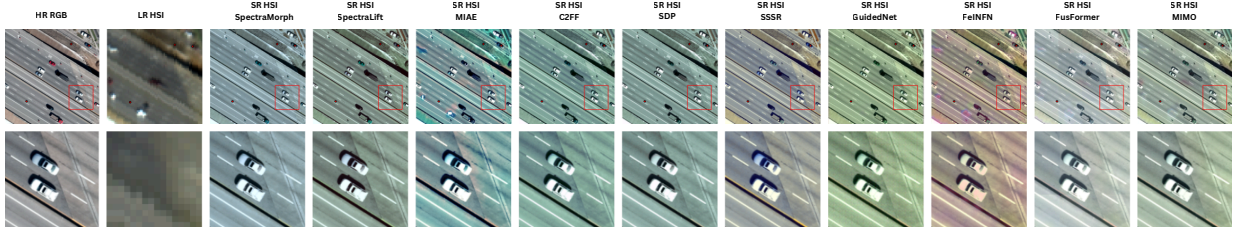}
    \caption{UH SR images for test scene 1. Second row shows a zoomed-in crop of the region marked in red.}
    \label{fig:uh_bottomright}
  \end{subfigure}
  \hfill
  \begin{subfigure}{\textwidth}
    \includegraphics[width=\textwidth]{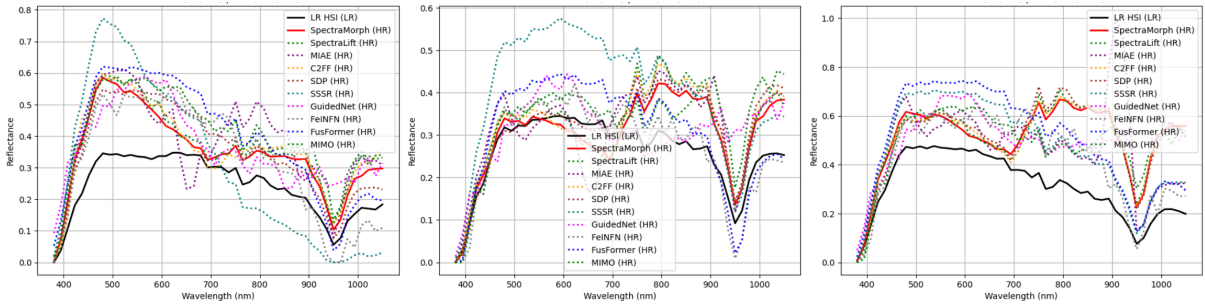}
    \caption{UH spectra for test scene 1. Left: Car, Center: Bare earth, Right: Highway.}
    \label{fig:uh_bottomright_spectra}
  \end{subfigure}

  \begin{subfigure}{\textwidth}
    \includegraphics[width=\textwidth]{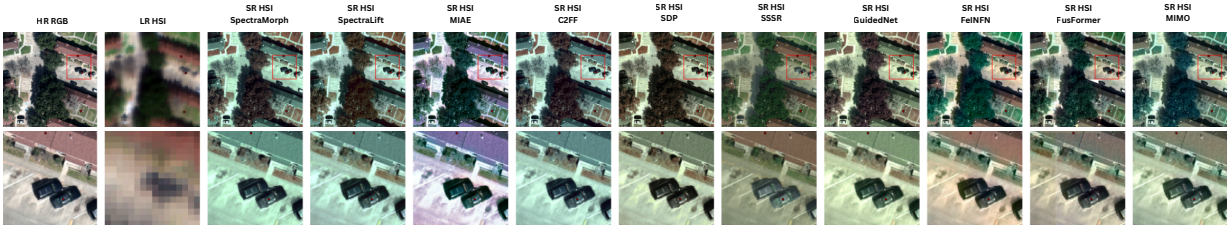}
    \caption{UH SR images for test scene 2. Second row shows a zoomed-in crop of the region marked in red.}
    \label{fig:uh_topleft}
  \end{subfigure}
  \hfill
  \begin{subfigure}{\textwidth}
    \includegraphics[width=\textwidth]{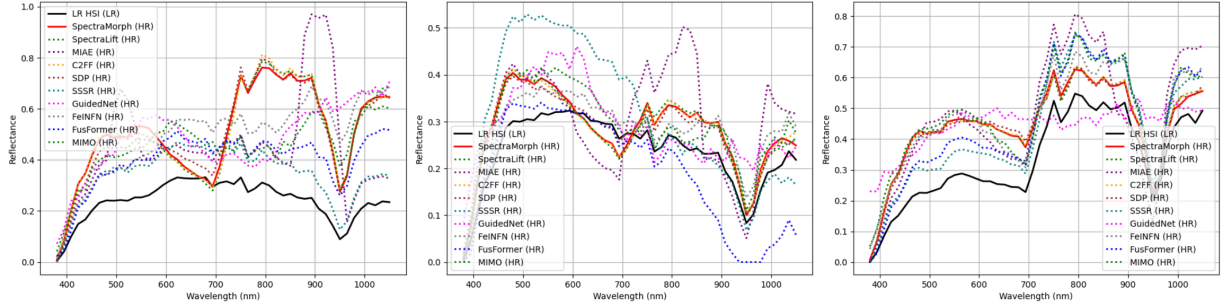}
    \caption{UH spectra for test scene 2. Left: Building roof, Center: Car, Right: Grass.}
    \label{fig:uh_topleft_spectra}
  \end{subfigure}

  \caption{UH SR results and corresponding spectra for two test scenes. (a, c) Super-resolved images with zoomed-in crops. (b, d) Spectral plots for selected regions.}
  \label{fig:uh_combined}
\end{figure*}

\subsection{Real-World Data}
We further evaluate SpectraMorph on the University of Houston (UH) dataset from the 2018 IEEE GRSS Data Fusion Contest~\cite{UH_data}, which provides an LR-HSI ($4172\times 1202\times 50$) and an HR-RGB ($83440\times 24040\times 3$). We remove the last 2 bands from the LR-HSI due to noise, resulting in an LR-HSI of shape ($4172\times 1202\times 48$). We extract two distinct $64\times 64\times 48$ regions from the LR-HSI and their aligned $1280\times 1280\times 3$ HR-RGB patches, downsample the RGB to $512\times 512\times 3$ ($r=8$), and apply all unsupervised methods (including SpectraMorph) to recover a $512\times 512\times 48$ HR-HSI. Testing is restricted to a small part of the UH dataset and the HR-RGB is downsampled from $1280\times 1280\times 3$ to $512\times 512\times 3$, in order to ensure that baseline methods could process the inputs in a single batch. If batch processing were used, the output of each batch would have to be stitched together, resulting in visible seam artifacts such as those seen for FusFormer in Fig.~\ref{fig:uh_bottomright} and~\ref{fig:uh_topleft}, which could not process each region in a single batch. SpectraMorph's pixel-wise processing is immune to such effects when inferred using batch processing.

Since UH lacks HR-HSI ground truth and its PSF/SRF are unknown, we use the IKONOS RGB SRF and let each unsupervised method estimate the PSF; due to our formulation, SpectraLift and SpectraMorph do not need to estimate the PSF. Supervised baselines were trained on Pavia Center HSI with inputs being generated by following Wald's protocol for $(r,c) = (8,3)$ via three overlapping $512\times 512\times 3$ HR-RGB patches and the corresponding LR-HSI patches, and then evaluated on the UH test pairs. For FusFormer, 53 non-overlapping HR-RGB patches of size $128\times 128\times 3$ and the corresponding LR-HSI patches had to be used to meet memory limits. 

It is important to note that we do not report proxy metrics such as Quality with No Reference (QNR) for the UH dataset. These metrics, while useful in certain controlled scenarios, are highly sensitive to the assumed PSF and SRF characteristics. In our case, no information regarding the PSF or exact SRF of the MSI acquisition is available, making such metrics unreliable and potentially misleading. Instead, we assess real-world performance through visual comparisons, generated by picking (Red = 659~nm, Green = 561~nm, Blue = 492~nm) bands from the HR-HSI outputs of each method (Fig.~\ref{fig:uh_bottomright},~\ref{fig:uh_topleft}) and spectral fidelity plots (Fig.~\ref{fig:uh_bottomright_spectra},~\ref{fig:uh_topleft_spectra}). 

Figure \ref{fig:uh_bottomright_spectra} plots the spectral signatures at three manually selected pixels -- car, bare earth, and highway -- marked in red in Figure \ref{fig:uh_bottomright}; Figure \ref{fig:uh_topleft_spectra} does the same for building roof, car, and grass pixels in Figure \ref{fig:uh_topleft}.  Because the spatial downsampling factor is $r=8$, each LR-HSI pixel $\Y_{i,j,:}$ corresponds to an $8\times8$ patch in the HR grid.  To compare spectra, we extract the spectrum of the LR pixel at $(i,j)$ and pair it with the spectrum of the HR-HSI at the top-right corner of its corresponding block, namely $(8i,\,8j)$.  For example, in Figures~\ref{fig:uh_bottomright_spectra} and~\ref{fig:uh_topleft_spectra} we plot LR-HSI pixel $(8,8)$ against HR-HSI pixel $(64,64)$ for a direct spectral fidelity comparison.

Because the proprietary SRF of the RGB camera used to capture the HR-RGB images is unavailable, we adopt the IKONOS RGB SRF in our experiments. However, the IKONOS SRF is likely a poor approximation of the true sensor response, which remains unknown and cannot be reliably estimated without manufacturer-provided specifications. In practice, the true SRF is always known to the camera manufacturer and would be available for real-world missions, eliminating this limitation. Unfortunately, the manufacturer has not disclosed the SRF of this particular camera, preventing a more accurate approximation. This substitution introduces color shifts in the HR-HSI outputs, particularly affecting unsupervised methods that explicitly rely on the SRF. Supervised methods are also influenced, as their synthetic training data uses the same IKONOS SRF to generate LR-HSI pairs. If the true SRF were available -- or if its characteristics (e.g., center, minimum, and maximum wavelengths of each band) were known, enabling a Gaussian or parametric approximation -- this limitation could be alleviated.

These tints are most visible in GuidedNet, FeINFN, FusFormer, and MIMO-SST in Figure~\ref{fig:uh_bottomright}, and in MIAE, SDP, SSSR, and GuidedNet in Figure~\ref{fig:uh_topleft}. Competing methods also exhibit other artifacts: GuidedNet yields subtle checkerboard patterns visible under magnification, and in Figure~\ref{fig:uh_bottomright}, where temporal misalignment between the LR-HSI and HR-RGB is most pronounced, methods that fuse the modalities jointly (MIAE, FeINFN, FusFormer, MIMO-SST) tend to produce ghosting and smearing around moving vehicles; by contrast, SpectraMorph is largely robust to this misalignment and maintains sharp boundaries aligned with the HR-RGB. That said, SpectraMorph is not artifact-free: it introduces a slight tint in Figure~\ref{fig:uh_topleft} and shows mild color inconsistencies in Figure~\ref{fig:uh_bottomright}. Among all methods, SpectraLift performs best in these color-sensitive regions, likely because it is free to directly optimize an SRF-inversion objective. In contrast, SpectraMorph is constrained to respect an endmember dictionary and is trained with ALLE to maintain that structure, which can bias colors slightly; this is worsened by the SRF mismatch.

Across both UH scenes, SpectraMorph recovers spectral shapes that closely match those of the LR-HSI, particularly for naturally occurring materials such as grass and bare earth (Figures~\ref{fig:uh_topleft_spectra}, \ref{fig:uh_bottomright_spectra}).  While absolute reflectance values deviate -- an expected consequence of the ill-posed spectral unmixing -- the overall spectral structure is faithfully preserved, supporting downstream analysis tasks that rely on relative band signatures. For certain man-made structures (highways and building rooftops), other baselines produce HR-HSIs whose spectra better aligns with that of the LR-HSI. 

\subsection{Oracle Analysis} 

Tables~\ref{tab:dc_quality} to~\ref{tab:botswana_quality} also report results for an \emph{oracle} version of SpectraMorph, which serves as a performance upper bound for our framework. Unlike the deployable setting, the oracle setup assumes access to high-resolution spectral information during training, and therefore cannot be realized in real-world applications. Its purpose is strictly diagnostic: to illustrate the best possible performance the model can achieve when provided with complete information.

Concretely, the oracle extracts the endmember matrix $\E$ directly from the entire ground-truth HR-HSI $\X$ using nonnegative matrix factorization (NMF). Training then follows the same supervised protocol as described in Section 3.3. The HR-MSI $\M$ is used as input, and optimization is carried out with the ALLE objective against the ground-truth HR-HSI. The training is performed on the 75\% crop of the HSI (same training setup as used for the comparison supervised baselines), while inference is conducted on the remaining 25\% crop that is mutually exclusive with the training set. Metrics are reported on this held-out region, ensuring a fair comparison with the deployable version of the model.

As expected, the oracle consistently outperforms  SpectraMorph across most datasets, since it benefits from HR-derived endmembers that more faithfully represent the underlying material spectra. The only exception is the KSC dataset, where oracle performance falls slightly below that of the deployable model. This is likely due to a distributional mismatch: certain endmembers present in the inference crop were not represented in the training portion, weakening reconstruction quality. The overall gap between the non-deployable oracle and the deployable SpectraMorph method remains modest. Although LR-based endmember extraction introduces some limitations, SpectraMorph remains self-supervised and highly competitive without access to HR spectra, and achieves performance that is close to the theoretical ceiling established by the oracle.

\subsection{Ablation}
\label{sec:ablation}

\begin{table*}[!t]
  \centering
  \captionsetup{font=small, justification=centering}
  \caption[Ablation study]{Ablation study of SpectraMorph on the Pavia Center dataset using only Gaussian PSF \\ (\textbf{best in bold}).}
  \label{tab:ablation_table}
  \resizebox{\columnwidth}{!}{%
  \begin{tabular}{|l|c|c|c|c|c|c|c|c|c|}
    \hline
    \textbf{Variant} & \textbf{RMSE $\downarrow$} & \textbf{PSNR $\uparrow$} & \textbf{SSIM $\uparrow$}
    & \textbf{UIQI $\uparrow$} & \textbf{ERGAS $\downarrow$} & \textbf{SAM $\downarrow$}
    & \textbf{Params (M) $\downarrow$} & \textbf{FLOPs (G) $\downarrow$} \\
    \hline
    Baseline & 0.0217 & 34.41 & \textbf{0.967} & \textbf{0.995} & \textbf{2.00} & \textbf{3.86} & 0.029 & 45.98 \\
    \hline
    Constraint: NN & 0.0689 & 27.62 & 0.881 & 0.905 & 5.52 & 12.40 & 0.029 & 45.98 \\
    \hline
    Constraint: NN and $\sum = 1$  & 0.0705 & 23.12 & 0.858 & 0.958 & 7.37 & 7.60 & 0.029 & 46.02 \\
    \hline
    No hidden layers in LEN & 0.0262 & 32.77 & 0.957 & 0.991 & 2.40 & 4.72 & \textbf{0.0002} & \textbf{1.70} \\
    \hline
    64 Neurons in LEN & 0.0226 & 34.00 & 0.965 & 0.994 & 2.09 & 4.04 & 0.002 & 4.23 \\
    \hline
    128 Neurons in LEN & 0.0226 & 34.10 & 0.965 & 0.994 & 2.06 & 3.96 & 0.004 & 7.01 \\
    \hline
    256 Neurons in LEN & 0.0221 & 34.30 & 0.966 & 0.994 & 2.03 & 3.91 & 0.007 & 12.58 \\
    \hline
    512 Neurons in LEN & 0.0218 & 34.36 & \textbf{0.967} & \textbf{0.995} & 2.01 & 3.87 & 0.014 & 23.71 \\
    \hline
    2048 Neurons in LEN & \textbf{0.0216} & \textbf{34.45} & \textbf{0.967} & \textbf{0.995} & \textbf{2.00} & \textbf{3.86} & 0.058 & 90.52 \\
    \hline
    2 hidden layers in LEN & 0.0218 & 34.40 & \textbf{0.967} & 0.994 & 2.01 & 3.89 & 1.079 & 1690.20 \\
    \hline
    MAE $\xrightarrow{}$ MSE loss & 0.0220 & 34.26 & 0.966 & 0.994 & 2.03 & 3.95 & 0.029 & 45.98 \\
    \hline
    ReLU $\xrightarrow{}$ Leaky ReLU & 0.0218 & 34.42 & \textbf{0.967} & 0.994 & \textbf{2.00} & 3.89 & 0.029 & 45.98 \\
    \hline
    ReLU $\xrightarrow{}$ GeLU & 0.0229 & 34.02 & 0.964 & 0.993 & 2.09 & 4.07 & 0.029 & 49.19 \\
    \hline
    ReLU $\xrightarrow{}$ tanh & 0.0231 & 33.98 & 0.964 & 0.994 & 2.10 & 4.08 & 0.029 & 45.98 \\
    \hline
    $K$ $\xrightarrow{}$ $K + 4$ & 0.0217 & 34.41 & \textbf{0.967} & \textbf{0.995} & \textbf{2.00} & 3.89 & 0.033 & 53.04 \\
    \hline
    $K$ $\xrightarrow{}$ $K - 4$ & 0.0227 & 33.90 & 0.966 & 0.994 & 2.10 & 4.07 & 0.025 & 38.92 \\
    \hline
  \end{tabular}}
\end{table*}

\begin{table*}[!t]
  \centering
  \footnotesize
  \setlength\tabcolsep{4pt}
  \renewcommand{\arraystretch}{1.1}
  \captionsetup{font=small, justification=centering}
  \caption{Ablation of coarse spectral prior: mean results for $(r,c)=(8,1)$ across all PSFs on all datasets \textbf{(best in bold)}.}
  \label{tab:csp_ablation}
  \resizebox{\columnwidth}{!}{%
  \begin{tabular}{|l|c|c|c|c|c|c|c|c|c|c|}
    \hline
    \textbf{Variant} & \textbf{RMSE $\downarrow$} & \textbf{PSNR $\uparrow$} & \textbf{SSIM $\uparrow$}
    & \textbf{UIQI $\uparrow$} & \textbf{ERGAS $\downarrow$} & \textbf{SAM $\downarrow$}
    & \textbf{Params (M) $\downarrow$} & \textbf{FLOPs (G) $\downarrow$}
    & \textbf{GPU Mem (MB) $\downarrow$} \\
    \hline
    Baseline ($s=4$) & 0.0548 & 25.81 & 0.860 & 0.967 & 5.75 & 6.84 & 0.1595 & 128.41 & \textbf{234.61} \\
    \hline
    Without CSP & 0.0822 & 22.76 & 0.811 & 0.929 & 8.00 & 10.97 & \textbf{0.0127} & \textbf{11.07} & \textbf{234.61} \\
    \hline
    ($s=2$) & \textbf{0.0537} & \textbf{25.93} & \textbf{0.864} & \textbf{0.968} & \textbf{5.72} & \textbf{6.75} & 0.1595 & 128.41 & \textbf{234.61} \\
    \hline
    ($s=8$) & 0.0579 & 25.42 & 0.850 & 0.958 & 5.97 & 7.19 & 0.1595 & 128.41 & \textbf{234.61} \\
    \hline
    ($s=16$) & 0.0673 & 24.21 & 0.833 & 0.946 & 6.88 & 8.53 & 0.1595 & 128.41 & 235.42 \\
    \hline
  \end{tabular}}
\end{table*}

To quantify the impact of our architectural and training choices, we evaluate a suite of SpectraMorph/LEN variants on the Pavia Center benchmark using only the Gaussian PSF; results are given in Table \ref{tab:ablation_table}. We perform the experiments using the same procedure as described in Section 3.3. All variants share the same hyperparameters and differ only in the component under test. Enforcing nonnegativity on the ALLE (Constraint: NN) via a ReLU activation on its output layer, or both nonnegativity and sum-to-one via a Softmax (Constraint: NN and $\sum = 1$), severely worsens all metrics. These constraints overly restrict the LEN, preventing it from flexibly fitting endmember mixtures. With respect to the LEN, the model performs competitively when its single hidden layer is reduced to just 64 neurons, with only small performance losses but significant computational complexity reduction as the number of neurons is reduced, highlighting that the unmixing bottleneck allows SpectraMorph to remain effective in resource-constrained settings. When using a linear head (removing the single hidden layer so the LEN only has an output layer) drastically worsens metrics, showcasing that non-linearity is necessary for this formulation. Increasing the complexity of the LEN's single hidden layer to 2048 neurons or introducing 2 hidden layers with 1024 neurons in each has a significant increase in computational complexity, which cannot be justified with the minimal metric gains. Swapping the MAE loss for MSE causes a small drop in fidelity. Changing activations -- ReLU to Leaky-ReLU, GeLU, or tanh -- has only a minor effect, showing robustness to the choice of nonlinearity. Moreover, increasing or decreasing the number of endmembers $K$ by 4-compared to the value of $K$ in the baseline setting-also had small effects on the metrics of the resulting super resolved output, showcasing the robustness of our choice of $K$. Nonetheless, a very poor selection of $K$ can result in extremely poor HSR. 

Table~\ref{tab:csp_ablation} evaluates the coarse spectral prior (CSP) in the panchromatic case, where \(s\) is the downsampling factor before block replication (Section~\ref{sec:coarse-prior}). Removing the CSP sharply degrades performance, while finer (\(s=2\)) and coarser (\(s=8\)) priors perform similarly to the default (\(s=4\)), showing that the main benefit comes from injecting spectral context rather than the exact choice of \(s\). However, increasing \(s\) results in the spectral signatures in $\V$ to become increasingly mixed due to spatial averaging. Hence, as \(s\) becomes too large (e.g., $s=16$) this mixing effect  reduces the much needed spectral context, leading to worsening performance as \(s\) increases beyond a certain threshold.

\section{Conclusion}
\label{sec:conclusion}
We presented SpectraMorph, a physics-guided self-supervised framework for HSR. The central idea is a structured unmixing bottleneck where an NMF-derived endmember dictionary is fixed and a compact latent estimation network predicts abundance-like representations without restrictive constraints. This design provides interpretable intermediates, eliminates the need for PSF modeling, remains agnostic to blur and input resolution, and enables training at a fraction of the cost of competing approaches. For the severely ill-posed panchromatic setting, we introduced a coarse spectral prior that stabilizes inversion by contributing only low-frequency spectral context.  

Extensive experiments on benchmark datasets show that SpectraMorph consistently achieves strong spatial and spectral reconstruction, outperforming existing unsupervised baselines and remaining competitive with supervised models that rely on HR ground truth. In addition, its lightweight formulation and robustness to temporal misregistration make it well suited for real-time or resource-constrained applications.  

\section*{Acknowledgement}
We thank Dr. Mario Parente for many helpful discussions on practical assessments of HSR methods.

\clearpage

\printbibliography

@article{wald,
  author    = {T. Ranchin and L. Wald},
  title     = {Fusion of high spatial and spectral resolution images: The {ARSIS} concept and its implementation},
  journal   = {Photogramm. Eng. Remote Sens.},
  volume    = {66},
  number    = {1},
  pages     = {49--61},
  month     = jan,
  year      = {2000}
}

@ARTICLE{MIAE,
  author={Liu, Jianjun and Wu, Zebin and Xiao, Liang and Wu, Xiao-Jun},
  journal={IEEE Trans. Geosci. Remote Sens.}, 
  title={Model Inspired Autoencoder for Unsupervised Hyperspectral Image Super-Resolution}, 
  year={2022},
  volume={60},
  number={},
  pages={1-12}}

@ARTICLE{C2FF,
  author={Li, Jiaxin and Zheng, Ke and Liu, Wengu and Li, Zhi and Yu, Haoyang and Ni, Li},
  journal={IEEE Geosci. Remote Sens. Letters}, 
  title={Model-Guided Coarse-to-Fine Fusion Network for Unsupervised Hyperspectral Image Super-Resolution}, 
  year={2023},
  volume={20},
  number={},
  pages={1-5}}

@ARTICLE{SDP,
  author={Liu, Jianjun and Wu, Zebin and Xiao, Liang},
  journal={IEEE Trans. Geosci. Remote Sens.}, 
  title={A Spectral Diffusion Prior for Unsupervised Hyperspectral Image Super-Resolution}, 
  year={2024},
  volume={62},
  number={},
  pages={1-13}}

@ARTICLE{GuidedNet,
  author={Ran, Ran and Deng, Liang-Jian and Jiang, Tai-Xiang and Hu, Jin-Fan and Chanussot, Jocelyn and Vivone, Gemine},
  journal={IEEE Trans. Cybern.}, 
  title={{GuidedNet}: A General {CNN} Fusion Framework via High-Resolution Guidance for Hyperspectral Image Super-Resolution}, 
  year={2023},
  volume={53},
  number={7},
  pages={4148-4161}}

@ARTICLE{FusFormer,
  author={Hu, Jin-Fan and Huang, Ting-Zhu and Deng, Liang-Jian and Dou, Hong-Xia and Hong, Danfeng and Vivone, Gemine},
  journal={IEEE Geosci. Remote Sens. Letters}, 
  title={{FusFormer}: A Transformer-Based Fusion Network for Hyperspectral Image Super-Resolution}, 
  year={2022},
  volume={19},
  number={},
  pages={1-5}}

@ARTICLE{MIMO,
  author={Fang, Jian and Yang, Jingxiang and Khader, Abdolraheem and Xiao, Liang},
  journal={IEEE Trans. Geosci. Remote Sens.}, 
  title={{MIMO-SST}: Multi-Input Multi-Output Spatial-Spectral Transformer for Hyperspectral and Multispectral Image Fusion}, 
  year={2024},
  volume={62},
  number={},
  pages={1-20}}

@ARTICLE{SSSR,
  author={Rajaei, Arash and Abiri, Ebrahim and Helfroush, Mohammad},
  journal={Sci. Rep.}, 
  title={Self-supervised spectral super-resolution for a fast hyperspectral and multispectral image fusion}, 
  year={2024},
  volume={14},
  number={1}}

@inproceedings{FeINFN,
 author = {Liang, Yu-Jie and Cao, Zihan and Deng, Shangqi and Dou, Hong-Xia and Deng, Liang-Jian},
 booktitle = {Adv. Neural Inf. Proc. Syst.},
 pages = {63441--63465},
 title = {Fourier-enhanced Implicit Neural Fusion Network for Multispectral and Hyperspectral Image Fusion},
 volume = {37},
 year = {2024}
}

@ARTICLE{UH_data,
  author={Xu, Yonghao and Du, Bo and Zhang, Liangpei and Cerra, Daniele and Pato, Miguel and Carmona, Emiliano and Prasad, Saurabh and Yokoya, Naoto and Hänsch, Ronny and Le Saux, Bertrand},
  journal={IEEE J. Sel. Top. Appl. Earth Obs. Remote Sens.}, 
  title={Advanced Multi-Sensor Optical Remote Sensing for Urban Land Use and Land Cover Classification: Outcome of the 2018 {IEEE GRSS Data Fusion Contest}}, 
  year={2019},
  volume={12},
  number={6},
  pages={1709-1724}}

@ARTICLE{SpectraLift,
      title={{SpectraLift}: Physics-Guided Spectral-Inversion Network for Self-Supervised Hyperspectral Image Super-Resolution}, 
      author={Ritik Shah and Marco F. Duarte},
      year={2025},
      journal={arXiv: 2507.13339},
      url={https://arxiv.org/abs/2507.13339}, 
}

@article{NNDSVD,
  title={{SVD} based initialization: A head start for nonnegative matrix factorization},
  author={Boutsidis, Christos and Gallopoulos, Efstratios},
  journal={Pattern recognition},
  volume={41},
  number={4},
  pages={1350--1362},
  year={2008},
  publisher={Elsevier}
}

\end{document}